# Boosting Brain-inspired Path Integration Efficiency via Learning-based Replication of Continuous Attractor Neurodynamics

Zhangyu Ge[†], Xu He[†], *Student Member, IEEE*, Lingfei Mo*, *Member, IEEE*, Xiaolin Meng*, Wenxuan Yin*, Student Member, IEEE*, Youdong Zhang, *Student Member, IEEE*, Lansong Jiang, Fengyuan Liu

***Abstract*—The brain's Path Integration (PI) mechanism offers profound insights for Brain-Inspired Navigation (BIN). However, numerous advances predominantly rely on Continuous Attractor Neural Networks (CANNs) to implement PI functionality, which introduces considerable computational redundancy and limits operational efficiency, thereby constraining the practicality of BIN technologies. To address this, this work proposes an efficient PI framework that leverages representation learning models to faithfully replicate the neurodynamic patterns of CANNs. Specifically, lightweight Artificial Neural Networks (ANNs) are employed to accurately emulate the neurodynamics of Head Direction Cells (HDCs) and Grid Cells (GCs) as modeled by CANNs. These AI-based HDC and GC representations are subsequently integrated to achieve brain-inspired PI for Dead Reckoning (DR). Benchmark evaluations across diverse environments, with comparisons against the established NeuroSLAM system, demonstrate that the proposed approach not only successfully replicates the neurodynamic behaviors of navigation-related cells but also achieves comparable positioning accuracy to NeuroSLAM. Furthermore, substantial efficiency gains are observed: approximately 17.5% improvement on general-purpose computing platforms and 40~50% on edge devices. This work thus presents a promising implementation strategy for enhancing the practicality of BIN technology and offers considerable potential for further development. GitHub link: https://github.com/BINUCOE/CANN2ANN_for_PI.**

## I. INTRODUCTION

In recent years, Brain-Inspired Navigation (BIN) research inspired by spatial cognition has attracted increasing attention and discussion within the research community. Current BIN research draws inspiration from the brain's navigational neural mechanisms, employing brain-inspired neurodynamic models such as Continuous Attractor Neural Networks (CANNs), etc. to realize navigation and positioning capabilities [1]. Given that BIN research is in its nascent stage, we cannot broadly and indiscriminately model and simulate all aspects of the brain's navigation circuitry. Therefore, many current studies focus on replicating key mechanisms within the brain's navigation-related neural circuits. Among these, the neural basis of Path Integration (PI) mechanism [2] provides crucial guidance for navigation tasks, such as Dead Reckoning (DR) and Simultaneous Localization and Mapping (SLAM).

A substantial body of existing research simulating the brain's PI capability adopts the CANN-based approach [3]. Diverging from conventional logic of numerical optimization-based uncertainty estimation, these approaches recast spatial state estimation as a property of internal state coupling and evolution within neurodynamic models, thereby reformulating spatial localization as an encoding-decoding process of spatial experiences [4]-[6].

Although this paradigm sacrifices some accuracy relative to computational methods, its operational principles bear closer affinity to the brain's construction of cognitive maps [7]. Consequently, it is regarded as a promising avenue for enhancing environmental adaptability under unstructured conditions [3][4]. This growing recognition has motivated considerable research effort toward exploring its scientific value, with particular significance lying in the potential to enable BIN systems to perform DR in unstructured complex environments, thereby substantially expanding the application scope of unmanned systems and intelligent robots.

While mechanistic modeling and computational paradigms have evolved considerably, CANN-modeled PI continues to present efficiency constraints that warrant attention. Recent studies by Zheng et al. [8] and Lu et al. [9] have highlighted that differential equation-based CANNs rely on cumbersome calculations to implement neurodynamics, motivating their simplified reformulations within probabilistic frameworks. Such adaptations, however, retain the core differential equation architecture rather than fundamentally restructure the computational workflow.

In fact, the theoretical foundations of CANNs were

This paragraph of the first footnote will contain the date on which you submitted your paper for review, which is populated by IEEE. This work was supported by the Natural Science Foundation of Jiangsu Province under Grant No. BK20243064. Xu He receives support from the Postgraduate Research & Practice Innovation Program of Jiangsu Province under Grant No. SJCX24_0067 and SEU Innovation Capability Enhancement Plan for Doctoral Students under Grant No. CXJH_SEU 24204.

*Xu He & Zhangyu Ge contributed equally and are co-first authors, with Xu He listed first alphabetically by surname. Corresponding author: Lingfei Mo & Xiaolin Meng.*

The authors are with the State Key Laboratory of Comprehensive PNT Network and Equipment Technology, School of Instrument Science and Engineering, Southeast University, Nanjing 210096, China (e-mail: lfmo@seu.edu.cn & xiaolin_meng@seu.edu.cn).

Xu He, Xiaolin Meng, Wenxuan Yin, Youdong Zhang, Lansong Jiang, and Fengyuan Liu are also with the China-UK Centre on Intelligent Mobility, Southeast University, Nanjing 210096, China.

established decades ago and have since found widespread engineering applications. Their enduring influence is evident in the lineage of brain-inspired SLAM systems, from the first-generation RatSLAM [10] in 2004, through NeuroSLAM [11] in 2019, to the recent ORB-NeuroSLAM [12] and many other advances like [13]-[16], all of which employ volumetrically stacked CANNs to construct core PI capabilities. This consistency attests to the CANN's neurobiological fidelity, yet also signals that its computational implementation has remained largely unchanged. The persistent reliance on iterative differential equation solvers and exhaustive decoding procedures, appropriate to early 2000s computational environments, now conflicts with the efficiency requirements of modern general-purpose and edge computing platforms in the era of Artificial Intelligence (AI).

Of course, it is understandable that as an emerging technology, the evolution of BIN in terms of its theoretical principles, computational paradigms, and functional implementations cannot be achieved overnight [3][4]. Its capability enhancement must be realized progressively. Therefore, at this critical stage, new ideas and methodologies can provide valuable contributions and insights to the research community of this emerging field. To this end, this paper investigates the optimization of computational efficiency in CANN-based modeling of neurodynamics.

Specifically, this paper employs AI-enabled techniques to replicate the neurodynamic patterns of CANN-modeled navigation cells into lightweight Artificial Neural Networks (ANNs). Then, they are encapsulated into a brain-inspired PI framework composed of Head Direction Cells (HDCs) and Grid Cells (GCs) to realize the PI capability. The objective is to enhance the practicality of brain-inspired PI methods by leveraging lightweight AI techniques, thereby accommodating the efficiency demands of modern edge computing and general-purpose platforms in the current AI era.

Furthermore, benchmark tests are conducted to verify the representational fidelity of this AI-replicated neurodynamics. Comparative evaluations against NeuroSLAM [11], a well-established system in the BIN community, demonstrate the validity of this work. Results indicate equivalent performance to the CANN-based NeuroSLAM system in dead reckoning tasks, while achieving operational efficiency gains of approximately 20% on workstation platforms and 40~50% at the edge, contingent upon dataset characteristics.

The *main contributions* are summarized as follows:

1) This paper proposes a learning-based approach to replicate the CANN's neurodynamics, enabling the use of lightweight ANNs as substitutes for CANN models. This facilitates the deployment of CANN-based models and their extended functions on general-purpose computing platforms or edge devices, allowing for more efficient operation.

2) This paper constructs the functional models of entorhinal cortex HDCs and GCs with the AI-replicated CANN neurodynamics and integrates them into a brain-inspired PI framework to support the implementation of PI functionality, enabling DR.

3) Through benchmark testing with the NeuroSLAM system, which is widely recognized within the BIN field, this work developed herein achieves consistent DR performance with the NeuroSLAM system while demonstrating notable efficiency improvements on general-purpose computing devices, especially at the edge.

*Outlines.* Section II presents the related work, Section III details the method design and comparison, Section IV covers the experiments and evaluation, Section V includes the discussion and analysis, and Section VI concludes the paper.

## II. Related Work

This paper primarily focuses on the replication of the neurodynamic patterns of CANNs into representation learning-based ANNs, as well as the reconstructed modeling of navigation cells for the integration into the brain-inspired PI framework. Given that Zhang et al. [17] have already reported foundational studies on the neurodynamic modeling of CANNs, this section mainly reviews the recent state-of-the-art advances in CANN-modeled navigation cells and PI methods.

### *A. CANN-based Modeling of Navigation Cells*

Neuroscience has shown that the brain's hippocampal-entorhinal cortex neural circuit contains a large number of heterogeneous navigation cells. Their interactions form the basis of spatial cognition and navigation abilities [18]. The PI capability emerging from the complex neural networks formed by these cells is a major focus of BIN research. Over time, to simulate this, the CANN models have been widely employed for modeling key navigation cells, such as pose cells, HDCs, GCs, and place cells [19]-[21], thereby laying the groundwork for brain-inspired PI.

Yu et al. [11], in reporting the NeuroSLAM system, provided a detailed presentation of the CANN-based modeling processes of HDCs and GCs in the entorhinal cortex, which is regarded as the key brain region for PI realization. This modeling method was inherited from the RatSLAM's approach of constructing CANN-based pose cells [6], including local excitation, global inhibition, activity normalization, cell state updating, and decoding. The CANN modeling approaches for other types of navigation cells are largely similar and can further achieve dimensional extension of navigation cells by altering the Gaussian distribution dimensions of the CANN model.

The above typical CANN-based navigation cell modeling approach can simulate the neural activation patterns of navigation neuron clusters, enabling one-to-one mapping between the PI states and the activity states of neuron clusters through decoding. Nevertheless, it exhibits considerable computational redundancy and computational complexity. As [22] stated, at any given time only about 1% of the cells in the CANN model corresponding to pose cells are activated, forming an envelope (i.e., the wave packet synthesis states of 3D CANN in various dimensions) that describes the system state. Moreover, the generation of this envelope follows the processes of local excitation, global inhibition, and activity normalization, each requiring iterative computation within a recurrent framework to establish the matrix states of CANN neuronal connection strengths. In addition, the exhaustive search for the center of the envelope during the state update and decoding processes of CANNs further exacerbates the problem of low computational efficiency.

The above issues motivated Zeng et al. [8] to develop a Bayesian probabilistic framework in the NeuroBayesSLAM system to simplify CANN modeling and thus construct a more efficient PI function. The Bayesian attractor model represents the network state through Gaussian distribution parameters, rather than adhering to the neuron activation patterns of local excitation and global inhibition. Although this parametric representation streamlines the computational process, it neither alleviates the inefficiencies inherent to the decoding procedure nor transcends the foundational reliance on differential equation-based neurodynamic modeling.

Therefore, based on the review and analysis of the above developments, this paper proposes to employ a non-stacked structure to model key navigation cells with low-dimensional CANN, and to replicate their neurodynamic properties through lightweight learning-based ANNs. On this basis, we can utilize end-to-end AI methods to completely reshape the modeling of neuronal activities and state decoding of navigation cells, thereby achieving high computational efficiency.

### *B. Advances in Brain-inspired PI Research*

Numerous studies have confirmed that GCs located in the medial entorhinal cortex are critically important for the realization of PI capability. However, in the early years, since neuroscience had not yet fully elucidated the contribution and role of the entorhinal cortex in spatial cognition, the first-generation RatSLAM [10] in 2004 abstracted virtual pose cells to perform PI functionality, based solely on simulating hippocampal place cells. In the RatSLAM system, Visual Odometry (VO) estimated self-motion cues. The pose cell model, responsible for PI, encoded and decoded these cues into spatial experiences. RatSLAM then achieved Loop Closure Detection (LCD) through visual template matching (local view cells) and converted discrete spatial experiences into a topological experience map.

At present, many derivative studies have been built upon the open-sourced OpenRatSLAM project [6], which significantly advanced the development of BIN research. Many subsequent studies, including BatSLAM [23], NeuroSLAM, and among others, have been derived almost entirely from RatSLAM, sharing the common logic of simulating the operational modes of navigation cells to replicate the brain's PI capability. In addition, some researchers have raised concerns that pose cells lack physiological evidence, potentially leading to limitations in adaptability and generalization [24].

Later, with advances in neuroscience, the mechanisms of spatial cognition in the entorhinal cortex were gradually uncovered, and the role of GCs was clarified. Many researchers began to simulate the multisource information processing mechanisms of the entorhinal cortex by compiling PI information separately through HDC models and GC models, thus moving beyond the paradigm of pose cell modeling. Representative studies consistent with these characteristics can be found in [11]-[16]. Subsequently, more mechanisms of navigation cells were simulated and integrated into the construction of brain-inspired PI capability. For example, Liu et al. [16] employed the CANN model to construct speed cell models and virtual distance cell models, which were combined with HDC and GC models to achieve brain-inspired PI capability in intelligent mobile applications. Moreover, some researchers have focused on the multiscale properties of navigation cells such as GCs, and on this basis developed PI models with multiscale navigation cell features, such as [14][15], which achieved certain progress.

However, it is not difficult to observe that although incorporating different functional types of navigation cells into the framework of brain-inspired PI construction provides certain capability improvements, it also significantly increases computational costs. These computational costs are positively correlated with the scale, number, dimensionality, and granularity of the navigation cell models built on CANN. Therefore, addressing the computational efficiency of CANN models is a worthy research direction, as it would enhance the practicality of brain-inspired PI models on general-purpose computing platforms and even on edge devices.

## III. Method Design

Figure 1 provides an overview of the efficient PI method proposed in this paper. The VO implementation adopted in this paper remains consistent with the method in NeuroSLAM [11], used only for estimating self-motion cues during the PI process, and thus no specific extensions are described.

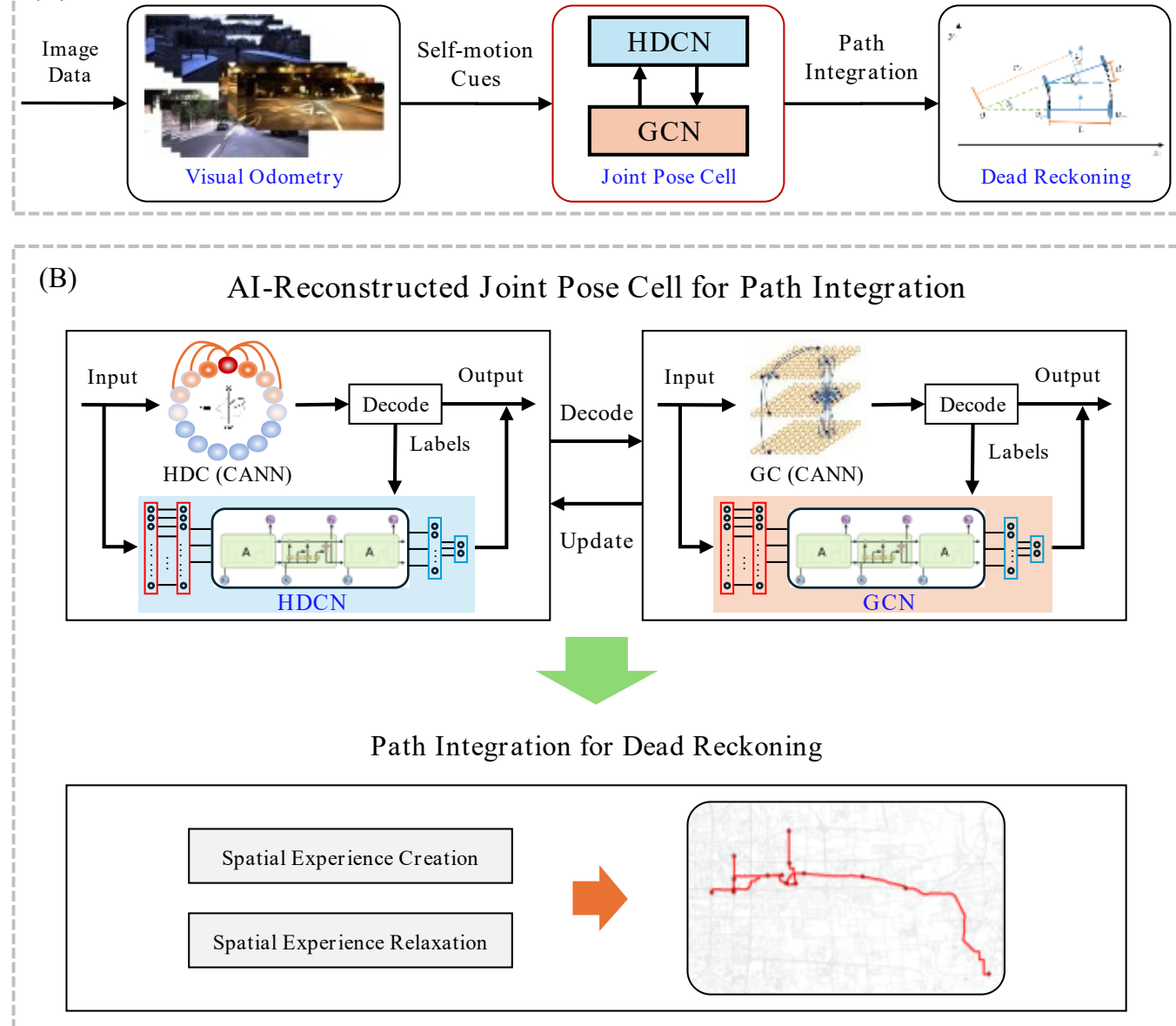

**Fig. 1.** Framework of the proposed PI method. (A) Pipeline overview. (B) Specific representation. Where, HDCN and GCN, representing Head Direction Cell Network and Grid Cell Network respectively, are detailed in the following subsections.

Since the open-sourcing of RatSLAM, a large number of brain-inspired PI models have adopted CANN models with stacked network structures. This structure causes each virtual neuron (cell) within it to form a subset representing the neurodynamic state. The superposition of these state subsets constitutes the complete envelope (or bump) of the CANN.

As mentioned before, this modeling approach faces significant computational redundancy during state updates of the CANN model, and its decoding efficiency is low. Its computational efficiency is positively correlated with the scale, number, dimensions, and size of the navigational cell model. Therefore, when we increase the scale, size, and quantity of the navigational cell model built with CANNs, the problem of low computational efficiency becomes more pronounced. To address this challenge, this work employs a learning-based approach to fully map the neurodynamic state update pattern and decoding process of the CANN. This effectively replicates CANN's neurodynamics, and performs a lightweight AI replication of the navigational cell patterns with enhanced computational efficiency, as illustrated in Figure 1 (B).

### *A. AI Replication of the CANN-modeled HDC*

By concatenating multiple 1D HDC models, we can achieve the encoding and decoding of yaw, pitch, and roll angles. This paper uses a multi-layer HDC model to estimate the yaw and pitch angles, temporarily disregarding roll angle information. This is consistent with Yu et al. [11], facilitating subsequent quantitative comparison in this paper. The difference is that, since the actual correlation between height and angle is not high, two tensors are used to maintain the hidden states of the temporal layers during cell reconstruction. This allows a single ANN fitting a 1D CANN to complete the calculation of the height and angle bumps, whereas the original algorithm effectively operates as a 2D CANN.

The design of Head Direction Cell Network (HDCN) for replicating the CANN-modeled multi-layer HDC patterns follows the lightweight principle, including an input layer, two Fully Connected (FC) layers with ReLU activation, three Long Short Term Memory (LSTM) layers, and two TimeDistributed layers for time-step state decoding. Among them, the two FC layers are used to extract abstract representations of the input sequence. To facilitate comparison with NeuroSLAM, both FC layers are set to 37 neurons. This is because the first and last neurons in the CANN model are completely equivalent, whereas in the NeuroSLAM's CANN modeling, the first and last neurons still have a spatial distance. This number setting ensures that the resolution of the AI-replicated HDCN remains consistent with the neuronal encoding resolution of the HDC model in NeuroSLAM. The three LSTM layers each have 37 Units, used to fit the neurodynamic patterns of the CANN; the two TimeDistributed layers have 12 neurons and two neurons respectively, used solely for decoding the head-direction patterns learned through representation learning by the LSTM layers.

HDCN is trained with supervised learning. The training dataset for HDCN is generated by a random process that creates sequential information about angles. The randomized elements include sequence length, the pattern of angle changes, the proportion of duration for different change patterns within the entire sequence, and the value when the angle is constant. The split ratio between the training set and the validation set is 4:1, and to reduce fitting difficulty and training instability, the sequence length is controlled to 500. The sequence length of the test set is 7-10 times that of the training set, not less than 3750 frames. The input values of the sequence vary linearly within $[-\pi, \pi]$ for different patterns or are a constant value, aiming to enable the network to fit wave packet activities under different rates of change. Before the input sequence is fed into the network, one-hot encoding plus interpolation is used to obtain the input vector for the network, allowing HDCN to preemptively capture the spatial proximity relationships between different inputs, thereby reducing the fitting difficulty. This paper defines a 37-dimensional vector $x \in \{0, 1\}^{37}$ corresponding to the input of HDCN, and specifies the preference vector for HDC as:

$$V_i = OHE(i) \tag{1}$$

Where, $i$ corresponds to the neuron index. Based on this, through interpolation, any input can be linearly mapped and transformed into a vector, as shown in formula (2):

$$V_{I_{ext}} = (1 - C_{frac}) OHE(C) + C_{frac} OHE(C+1) \tag{2}$$

Where, $C = I_{ext}/2\pi * N_P$, $C_{frac}$ refers to the fractional part of $C$, and $N_P$ is the number of intervals between neurons in the 1D CANN. The external input $(I_{ext})$ is constrained to $[0, 2\pi)$ after a modulo operation.

The training details for HDCN are described as follows. This paper uses the decoded values of the CANN-modeled HDC as the training labels for the model. For the involved decoding methods, see formulas (7) and (8). During HDCN training, the Adam optimizer and the Mean Squared Error (MSE) loss function were adopted, training for a total of 200 epochs, with 12 sequences input per epoch.

The CANN modeling of the HDC neurodynamic pattern to be fitted is described as follows:

$$I_{ext}^i = a_{ext} \exp\left[ -b_{ext} \left\| p_{ext} - p_i \right\|^2 \right] \tag{3}$$

Where, $I_{ext}^i$ represents the stimulus input to the $i$ th neuron. $a_{ext}$ and $b_{ext}$ are constants greater than 0. $p_{ext}$ and $p_i$ are the preferences of the external input and the $i$ th neuron, respectively, both taken from the information state set. For example, when constructing HDC via CANN, the preference here is a certain angle within $[-\pi, \pi]$. When constructing GC, it refers to a coordinate within a specified spatial range. At this point, the neuron state update equation and the internal connection weights can be described as:

$$\tau \frac{\partial U^i(t)}{\partial t} = -U^i(t) + \rho \sum_j J_{ij} r^j(t) + I_{ext}^i(t) \tag{4}$$

Where, $\tau$ and $\rho$ represent the synaptic time constant and neuron density, respectively. $J_{ij}$ denotes the connection weight between the $i$ th and $j$ th neurons. $r^j(t)$ represents the firing rate of the $j$ th neuron. The synaptic input $U^i(t)$ for the $i$ th neuron can be calculated using formula (4).

$$J_{ij} = \frac{J_0}{\sqrt{2\pi} a} \exp\left[ \frac{-\left\| p_i - p_j \right\|}{2a^2} \right] \tag{5}$$

The connection weights between neurons are initialized using the Gaussian function in formula (5), utilizing parameter $a$ to control the connection range of the neurons.

$$r^i(t) = \frac{\left[ U^i(t) \right]_+^2}{1 + k\rho \sum_j \left[ U^j(t) \right]_+^2} \tag{6}$$

Where, $0 < k < k_c \equiv \rho J_0^2 / 8\sqrt{2\pi} a$, $[U]_+ \equiv max(U, 0)$.

The divisive normalization in formula (6) limits the intensity of the neuronal firing rate, thereby preventing the synaptic input $U(t)$ from exhibiting an explosive growth trend. Formulas (3)-(6) complete the entire process from CANN receiving external input to wave packet state update.

In practical BIN applications, information transfer between different modules cannot be directly transmitted through abstract neural information but relies on paradigms of mathematical calculation. Therefore, it is necessary to decode the wave packet state of the CANN to obtain concrete information. For the wave packet state of a 1D CANN, its decoding formulas are written as follows:

$$\theta(t) = \arctan 2\left(\sum_i U^i(t)\sin\left(\frac{2\pi\cdot p_i}{N}\right), \sum_i U^i(t)\cos\left(\frac{2\pi\cdot p_i}{N}\right)\right) \tag{7}$$

$$y(t) = \left(\frac{\theta(t)\cdot N}{2\pi}\right) \bullet \mathrm{mod}(Z) \tag{8}$$

Where, $Z$ is the number of neurons in the CANN, and $y$ represents the decoding result of the current wave packet.

### *B. AI Replication of the CANN-modeled GC*

Similarly, this paper extends the 1D CANN-modeled HDC model into a 3D CANN-modeled GC model. By constructing a 3D CANN implementation, the encoding capability of GC for 3D coordinate information can be simulated. At this point, the wave packet generated by neuronal activity in the CANN-modeled GC also transitions from the 1D ring structure of HDC to a spherical structure in 3D space. Integrating the PI update rules and decoding this spherical wave packet can simulate the GC working mode [12]. The modeling of the 3D CANN can be directly extended from the 1D CANN model. However, for a K-dimensional CANN model, assuming each dimension has N neurons, its time and space complexity are both $O(N^K)$, making the complexity of fitting the neurodynamic patterns of a 3D CANN model significantly higher than that of a 1D ones.

The design of Grid Cell Network (GCN) for replicating the CANN-modeled 3D GC patterns is the same as that of HDCN, including an Input layer, two ReLU-activated FC layers, three LSTM layers, and two TimeDistributed layers for time-step state decoding. Due to the complexity issues mentioned earlier, for multi-dimensional CANN models, this paper does not directly fit the activity states of the neurons but instead fit the vectors after dimensionality reduction and concatenation during the decoding process. The dimensionality reduction method will be introduced later. This strategy reduces both time and space complexity to $O(N \cdot K)$, which is also the key to the improved running speed of this work, and the pre-reduction wave packet state can be approximately obtained through inverse transformation.

To ensure that the resolution of GCN is consistent with HDCN to match the original NeuroSLAM settings, the number of neurons in the FC and LSTM layers of GCN are set to 111, used to replicate the neurodynamic patterns of the CANN-modeled GC. The two TimeDistributed layers have 37 and 6 neurons, respectively, with every two output neurons corresponding to one dimension.

The dataset used for supervised training is also generated through a random process, identical to the random process used to generate input data for training HDCN. When creating the input sequence dataset, three 1D random sequences are pre-generated and then concatenated to simulate coordinate changes. The range of input values is still controlled within $[-\pi, \pi]$, and GCs of different scales can be mapped into this range. During input data preprocessing, this paper concatenates the input vectors calculated for the 3D coordinates according to formula (9).

$$V_I^T = \left[V_{x_I}^T, V_{y_I}^T, V_{z_I}^T\right] \tag{9}$$

The modeling process for GC is similar to that of HDC, with the difference being that the neuron preference changes from 1D angle data to a 3D vector describing coordinates. Its modeling process is shown in formulas (10) and (11):

$$\tau\frac{\partial U^i(t)}{\partial t} = -U^i(t) + \rho\sum\nolimits_x\sum\nolimits_y\sum\nolimits_z J^i_{x,y,z} r^i(t) + I^i_{ext}(t) \tag{10}$$

$$J^i_{x,y,z} = \frac{J_0}{\sqrt{2\pi}a}\exp\left[\frac{-\left\|p_{x_i,y_i,z_i} - p_{x,y,z}\right\|}{2a^2}\right] \tag{11}$$

Where, $J^i_{x,y,z}$ represents the connection weight between the $i$th neuron and the neuron at spatial coordinate $(x, y, z)$. $(x_i, y_i, z_i)$ represents the spatial coordinate of the $i$th neuron. $p_{x,y,z}$ is the preference of the neuron at spatial coordinate $(x, y, z)$. Here, $I^i_{ext}$ is the weighted input obtained through Gaussian encoding, and its calculation method is similar to Equation (3). Therefore, the right-hand side of the state update formula can be regarded as $-U^i(t) + F(x_i, y_i, z_i)$, and the three inputs of $F(x_i, y_i, z_i)$ have independent effects on the mapping results. Hence, from the perspective of the fitting target, the state update process of the original CANN model is also independent across the three dimensions and does not possess correlation.

The encoding and normalization process for the input sequence remains consistent with the description in HDCN. When decoding the GC wave packet, it is necessary to first sum the cell activity array according to dimensions to obtain 1D neuronal information representations on the X, Y, and Z dimensions respectively, as shown in formula (12):

$$\begin{aligned} SX(x) &= \sum\nolimits_y\sum\nolimits_z U^{x,y,z}, \\ SY(y) &= \sum\nolimits_x\sum\nolimits_z U^{x,y,z}, \\ SZ(z) &= \sum\nolimits_x\sum\nolimits_y U^{x,y,z} \end{aligned} \tag{12}$$

Where, $SX$, $SY$, $SZ$ represent the 1D vectors reduced to the $x$, $y$, $z$ dimensions, respectively. Then, the decoded values for each dimension are calculated to obtain spatial information. Here, the concatenated result of $SX$, $SY$, $SZ$ is the fitting target for the LSTM layer of GCN.

The training details for GCN are described as follows. This paper uses the decoded vectors from the CANN-simulated GC as the training labels for the model. These decoded vectors are concatenated from the numerator and denominator values corresponding to the arctangent calculation during the decoding of $SX$, $SY$ and $SZ$ respectively. The GCN training process is identical to the HDCN training process and will not be repeated here.

### *C. Relaxed Association of Spatial Experiences for DR*

Based on the reconstruction of HDC and GC, HDCN and GCN can directly decode the spatial experiences corresponding to the self-motion cues from the original VO input through mutual collaboration. However, it is still necessary to consider the geometric relationships between these discrete spatial experiences to form a complete PI logic, enabling the basic capability of DR, as illustrated in Figure 2.

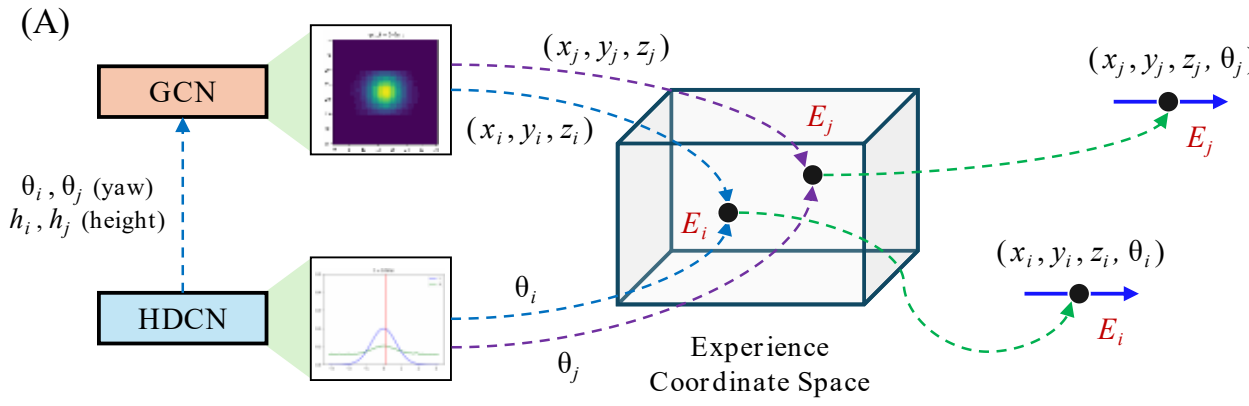

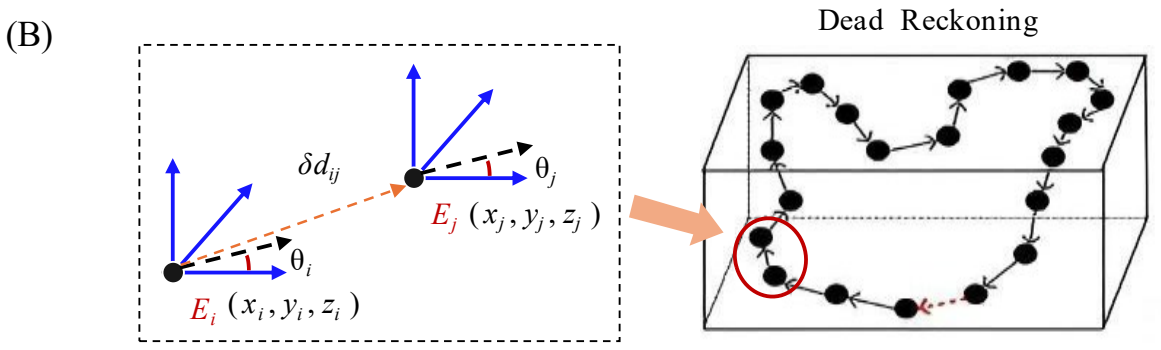

**Fig. 2.** Illustration of relaxed association of spatial experiences decoded from the PI framework for DR.

A single spatial experience node is defined as $E_i$, represented by the set containing the decoded state of the joint pose estimation cells and the current pose state from VO, see (13).

$$E_i = \left\{ P_i^{gc}, P_i^{hdc}, P_i^{exp} \right\} \tag{13}$$

Where, $P_i^{gc}$ and $P_i^{hdc}$ represent the decoded states of GCN and HDCN in the joint pose cells for the $i$ th spatial experience, respectively, and $P_i^{exp}$ represents the pose representation of the $i$th spatial experience estimated from the self-motion cues of VO. Since this paper follows the same VO implementation scheme as NeuroSLAM, the definition and connotation of $P_i^{exp}$ are consistent with it. Details can be found in [11] and are not repeated here. Associating the joint pose estimates provided by $P_i^{gc}$ and $P_i^{hdc}$ with the $P_i^{exp}$ provided by VO allows the spatial experience node to correspond to a discrete spatial state $(x, y, z, \theta)$.

The pose of the initial spatial experience node is defined as the zero initial state. To generate a new spatial experience node, the current node must be significantly different from all existing nodes. For convenience, this paper defines $P^{gc}$ and $P^{hdc}$ as the decoded states concerning GCN and HDCN in the joint pose cells among all existing nodes.

This difference evaluation metric can be determined by the metric score described in formula (14).

$$S_i = \mu^{gc} \left| P_i^{gc} - P^{gc} \right| + \mu^{hdc} \left| P_i^{hdc} - P^{hdc} \right| \tag{14}$$

Where, $\mu^{gc}$ and $\mu^{hdc}$ represent the weight contributions of GCN and HDCN, respectively. If all matching scores are below a set threshold, the spatial experience node corresponding to the lowest score (i.e., the $E_i$ corresponding to the index of $min\, S_i$) is activated, representing the current PI state of the BIN system. If the current metric score exceeds a preset threshold, a new spatial experience node is generated along with a transition link associated with it. To explain this process conveniently, this paper has drawn Figure 2 (B), showing the association logic from an existing spatial experience node $E_i$ to a newly added spatial experience node $E_j$. This transition link stores the incremental information of the pose change $\delta P_{ij}^{exp}$, as well as the distance $\delta d_{ij}$ between the two connected spatial experience nodes (i.e., the Euclidean distance of the $\delta P_{ij}^{exp}$ vector).

Specifically, the calculation rule for $\delta P_{ij}^{exp}$ is described as follows:

$$\delta P_{ij}^{exp} = P_j^{exp} - P_i^{exp} \tag{15}$$

On this basis, the newly added spatial experience node $E_j$ can be represented as:

$$E_j = \left\{ P_j^{gc}, P_j^{hdc}, P_i^{exp} + \delta P_{ij}^{exp} \right\} \tag{16}$$

The above spatial experience creation and update rules are the prerequisite for applying relaxation processing to all spatial experience nodes. On this basis, this paper adopts the following calculation rule to correct the pose estimate corresponding to the $i$ th spatial experience node $E_i$.

$$\delta P_i^{\exp} = \alpha \left[ \sum_{k=1}^{N_k} \left( P_k^{\exp} - P_i^{\exp} - \delta P_{ki}^{\exp} \right) + \sum_{t=1}^{N_t} \left( P_t^{\exp} - P_i^{\exp} - \delta P_{ti}^{\exp} \right) \right] \tag{17}$$

Where, $\alpha$ is the correction rate, $N_k$ and $N_t$ represent the number of transitions from $E_i$ to other spatial experience nodes, and the number of transitions from other spatial experience nodes to $E_i$, respectively. On this basis, integrating the above pose correction rule into the relaxation correction iteration process allows for the update of the spatial states $(x_i, y_i, z_i, \theta_i)$ corresponding to all discrete spatial experiences, forming complete PI information interpretation and generating the DR trajectory.

## IV. Experiment and Results

To demonstrate the effectiveness of this work, experiments were conducted. NeuroSLAM was set as the baseline during the whole evaluation processes. The aim of this evaluation is to demonstrate that this work can not only reliably replicate the neurodynamic patterns of navigation cells modeled by CANNs with different dimensions, but also achieve equally instead performance to NeuroSLAM's PI framework to perform DR. Furthermore, real-world testing was carried out in this paper. Evaluations demonstrated the computational efficiency advantages of the proposed efficient PI method. These evaluations were conducted on a Personal Computer (PC) and an edge device to showcase its practicality and transferability.

### *A. Data Preparation*

The first dataset sources from Yu et al.'s NeuroSLAM work [11]. It was collected in a two-story parking garage environment, encompassing both indoor and outdoor scenes. It consists of 12583 color images with a resolution of 480*270 pixels, covering a total travel distance of approximately 600 meters. However, it lacks Ground Truth (GT) annotation. More details can be found in the original publication. The parameter configuration used on this dataset was identical to that used by Yu et al., to validate the effectiveness of the AI-replicated neurodynamic representations of navigation cells. This was done to evaluate whether this work can fully substitute the PI capability of the NeuroSLAM based on CANN modeling.

Additionally, this study collected a self-built dataset, described as follows. Representative environmental views corresponding to this dataset are shown in Figure 3. The environment is a long, narrow circular corridor with a flat floor and vertical walls. Its structure is relatively monotonous, exhibiting numerous repetitive visual features and textures with low distinctiveness. During the data collection process, pedestrians occasionally appeared suddenly in the environment, posing additional challenges to the benchmarking.

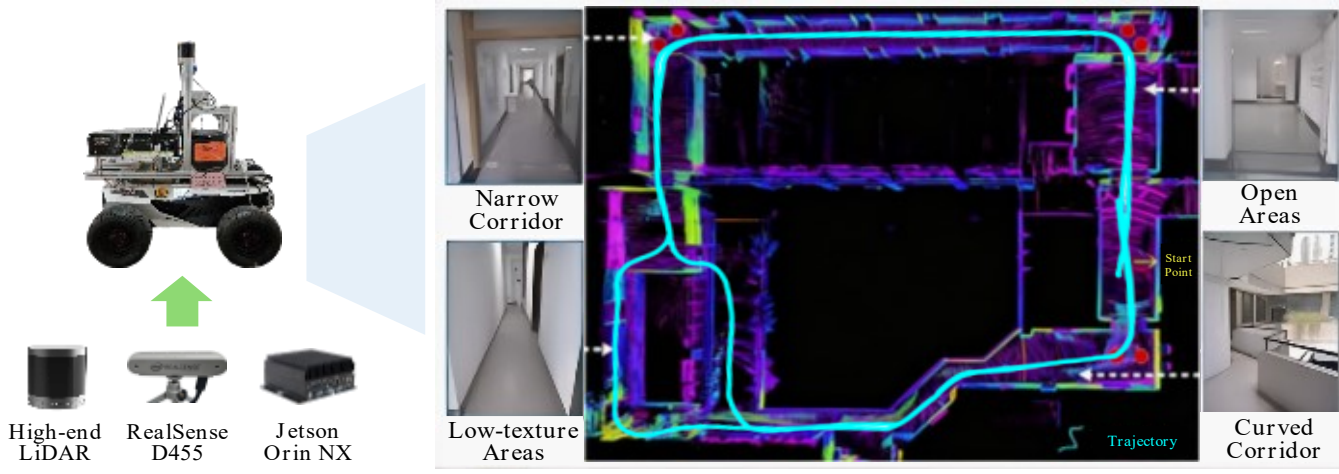

**Fig. 3.** Data collection environment and calibration results for the self-built dataset.

### *B. Experimental Conditions*

Experiments in this paper were conducted on both PC and edge platforms. Training of the relevant ANN models was performed on a PC equipped with one NVIDIA RTX 4060 GPU (8GB RAM) and one Intel Core i7-12650H CPU (2.7 GHz). Training data were generated using the BrainPy framework [25]. The PC's software environment was configured with PyTorch 2.5.1 and Python 3.9. The edge device is a Raspberry Pi 4B (8GB RAM). Its system environment was Raspbian GNU/Linux, and the software environment was configured with Keras 2.3.1 and Python 3.7.

When evaluating on the first dataset (two-story parking garage), the parameters involved were kept consistent with the original work. The parameter configuration used on the custom dataset was tuned via an optimizer. Parameter configurations, except those related to navigation cell modeling, were shared between the original NeuroSLAM system and the proposed efficient PI method. Parameters related to navigation cell modeling remained consistent across different datasets. This approach ensures fairness in the benchmark tests.

Since the AI-replicated HDCN and GCN in this study no longer rely on Gaussian functions to construct the neurodynamic characteristics of CANN (such as local excitation, global inhibition, and state update), there are no corresponding parameter settings to specify. Furthermore, the parameter settings for the NeuroSLAM system used as the baseline in this paper are consistent with the configurations used in [11] and [12]. Relevant descriptions can be found in the parameter definitions and accompanying explanations provided in [11] and [12].

### *C. Experimental Results of PI*

NeuroSLAM does not strictly build its experience map frame by frame, but rather constructs the map based on spatial information representation differences at different locations. Consequently, there is only topological similarity between its experience map and the VO trajectory, preventing direct calculation of translational error against a GT trajectory. To enable error calculation, this study tracked the visual template update information recorded by NeuroSLAM to identify the image frames corresponding to experience map locations. This allowed for sampling and alignment of the map trajectory to compute the Mean Absolute Error (MAE) and Root Mean Square Error (RMSE).

**Experiment 1:** Since the first dataset lacks GT annotations, the evaluation primarily aimed to demonstrate that this work can fully replicate the neurodynamic representations of CANN-modeled PI capabilities of NeuroSLAM system.

Figure 4 visually compares the accuracy of the experience maps generated by this work and NeuroSLAM. The relevant parameter configurations have been described previously. The DR result using pure VO showed severe trajectory drift due to accumulated error after multiple loops, whereas NeuroSLAM avoided this issue and produced a stable topological experience map. According to Figure 4, the DR trajectories generated by this work and NeuroSLAM are almost completely coincident. This observation proves that this work effectively replicated the neurodynamic representations of the CANN-modeled navigation cells, achieving functional equivalence.

Moreover, compared to NeuroSLAM, the operational efficiency of this work improved by 17.5% on the PC platform. For fairness, the LCD module was excluded from NeuroSLAM during time-cost testing. Similarly, on the Raspberry Pi 4B, the operational efficiency improved by approximately 39.4%. Additionally, the involved algorithms were not quantized or subjected to computational precision truncation on either PC or a Raspberry Pi. Therefore, their accuracy remains consistent across platforms/systems, and this will not be reiterated below to avoid redundancy.

**Experiment 2:** The following records the test conducted using the self-built dataset. Since this dataset possesses GT annotations with a confidence accuracy of 10 cm error. Therefore, Experiment 2 involved quantitative accuracy assessment of various methods. The GT trajectory for this dataset is shown in Figure 3. This represents a typical benchmark test with challenging settings, which is entirely different from the scenario faced in the benchmark test on the first dataset in Experiment 1.

As shown in Figure 5, the pure VO's trajectory did not form a closed loop. Since the test environment of Experiment 2 is rather challenging, so pure-VO drift becomes extremely severe, especially in the corridor sequence of this dataset (bottom-left panel of Figure 3's GT recording), with narrow space, low light and weak texture natures.

In contrast, the DR results from both this work and NeuroSLAM produced closed trajectories, showing much better geometric consistency with the GT trajectory. Yet, due to the severe VO drift, both NeuroSLAM and the proposed PI approach can only guarantee meter-level accuracy. However, the performance of this work and NeuroSLAM in the new test environment remained almost identical, further supporting its validity.

To ensure temporal alignment among the trajectories from different strategies, all error metrics are computed on the down-sampled sequences. Table I records the key metrics of Experiment 2. The travelled distance of the GT trajectory is 126.4 m. Under this setting, the proposed method achieves a MAE and RMSE of the translational drift of 2.0 % and 2.3 %, respectively. The Maximum Trajectory Error (MTE) of the experience map is about 5.9 m, outperforming the VO counterpart (12.3 m). Runtime evaluations further reveal that the proposed implementation is more efficient, reducing the total execution time by 17.5 % on the PC and by 48.9% on the Raspberry Pi 4B.

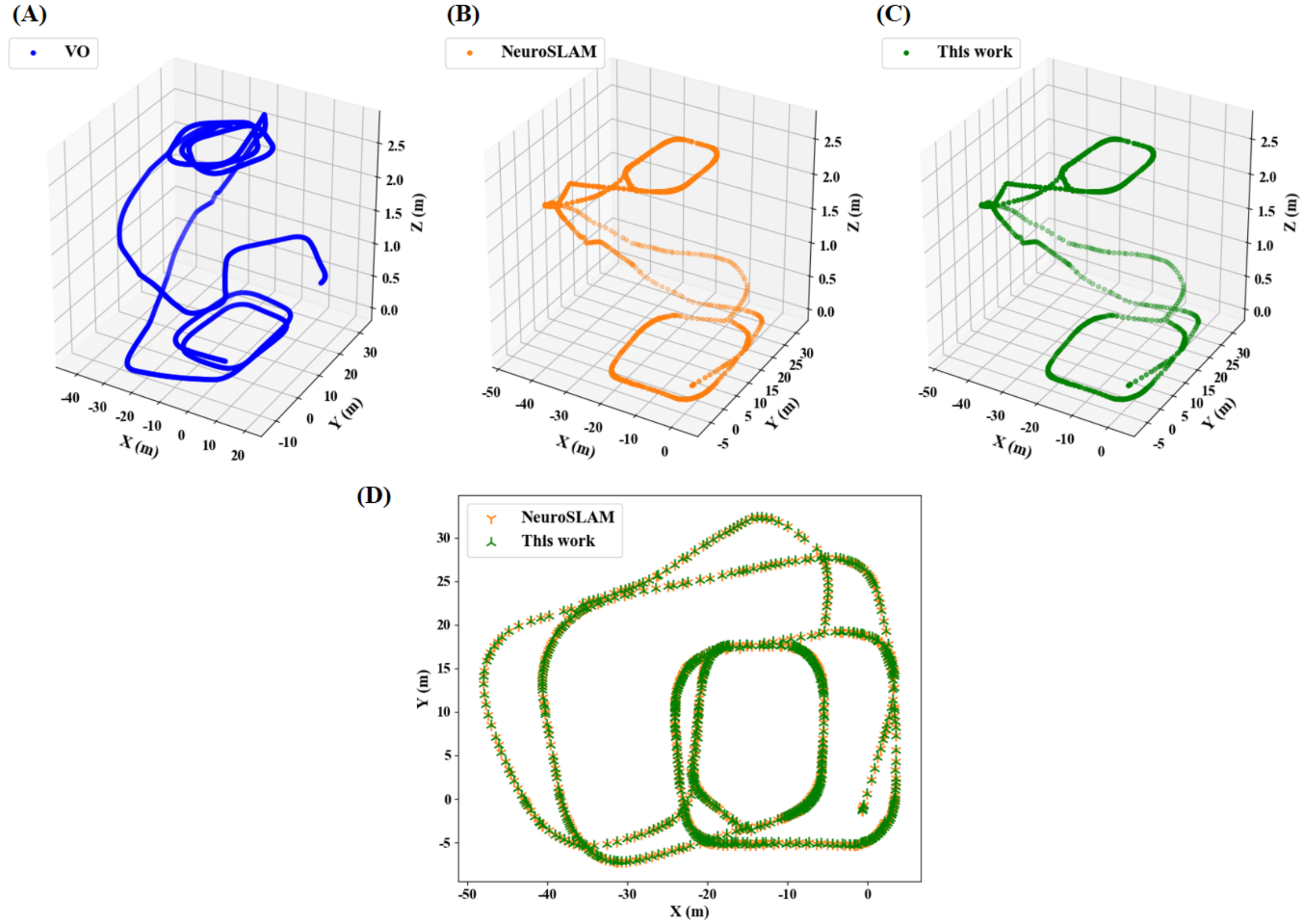

**Fig. 4.** Accuracy comparison (Experiment 1). (A) DR trajectory from pure VO. (B) DR trajectory from NeuroSLAM. (C) DR trajectory from this work. (D) Top-down view comparison within the same reference frame.

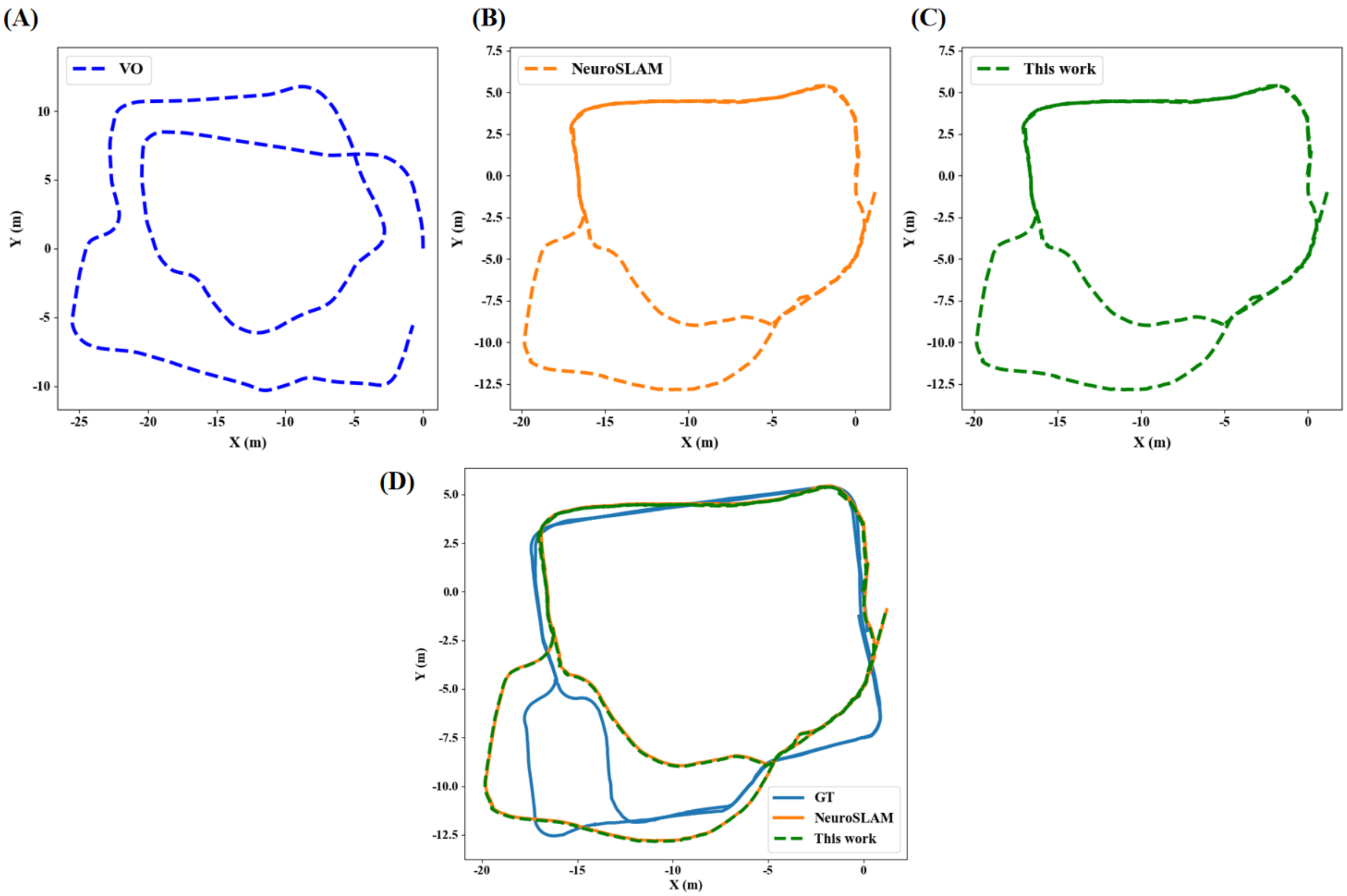

**Fig. 5.** Accuracy comparison (Experiment 2). (A) DR trajectory from pure VO. (B) DR trajectory from NeuroSLAM. (C) DR trajectory from this work. (D) Top-down view comparison within the same reference frame.

TABLE I
TRAJECTORY PERFORMANCE COMPARISON

| | Frame | Resolution | Time Cost (s) | MAE (m) | RMSE (m) | MTE (m) |
|---|---|---|---|---|---|---|
| Pure VO [6] | | | -- | 7.0 | 7.5 | 12.3 |
| NeuroSLAM [11] | 2613 | 320*180 | 871.4 @ PC<br>1478.1 @ Edge | 2.5 | 2.9 | 5.9 |
| This work | | | 721.6 @ PC<br>756.1 @ Edge | 2.5 | 2.9 | 5.9 |

In addition, this paper modified the HDCN and GCN network resolutions for further performance evaluations. We trained two additional HDCN and GCN networks with resolutions of 25/75 and 49/147 neurons, compared to the original 37/111 configuration tested above. Tested on the Raspberry Pi 4B using the self-built dataset, repeated tests show execution times for each hyperparameter configuration consistently range from 730s to 750s, with minimal variation (about 2%).

All the three achieved identical trajectory error performance. The identical trajectory accuracy across all hyperparameter configurations likely stems from the sufficiently redundant minimum resolution of the CANN and ANN used for spatial experience encoding. Moreover, minor changes in ANN parameters under different hyperparameter fluctuations do not significantly impact inference efficiency on the Raspberry Pi 4B. Similarly, during additional PC tests, inference times remained very close across all hyperparameter configurations, making a special explanation unnecessary.

*D. Evaluations on AI-replicated Neurodynamics*

The above subsection has demonstrated the quantitative evaluation metrics of this work in practical environments and benchmark tests, proving the engineering practical value of the proposed method. This section supplements the functional tests by analyzing the neural activity patterns of the navigation cells. It aims to evaluate the fitting effect of AI-replicated CANN neurodynamics.

This work differs from [26] and [27], which use Recurrent Neural Networks (RNNs) and Spiking Neural Networks (SNNs) to fit the relationship between self-motion cues and real-world navigation data with neural constraints, relying on real-world GT annotations. Our work aims to replicate CANN-modeled navigation cells using ANNs, focusing on achieving identical encoding-decoding capabilities within a brain-inspired PI framework. Since CANN's attractor characteristic involves one-to-one mapping of input sequences to state space subsets, capturing its mapping logic ensures functional equivalence. Thus, training ANNs to fit CANNs in this work does not require real-world navigation data. Instead, comprehensive simulated data covering ideal conditions, noise, and nonlinearity is needed to fully replicate CANN's neurodynamic characteristics. Moreover, we have considered the potential impact of instantaneous acceleration when generating synthetic data and introduced some abrupt situations, which are more realistic to actual motions, into the dataset, but not as the main part.

Figures 6 and 7 show the performance of the fitted 1D CANN and 3D CANN considering nonlinear motion conditions, respectively. The generation function used for the input sequences in the figures is exactly the same as the one used in the training set. It can be seen that the inputs have a certain degree of dynamics and nonlinearity, with multiple abrupt changes and pauses, and the rate of change is not entirely consistent at different moments. Judging from the decoding effects in Figures 6 and 7, during long-term operation of more than 4,800 frames, the decoding effect fitted by ANNs does not deviate significantly from that of CANN, when facing several abrupt changes. Therefore, it can be concluded that the AI-replicated CANNs have sufficient robustness against instantaneous changes.

It is worth mentioning that this robustness is demonstrated not only on the synthetic dataset but also in real-world verification. We considered the impact of instantaneous changes and validated this robustness practically. Experiment 2 featured significant nonlinear motions, including instantaneous start-up, sharp turns, vertical jolts, lighting changes, and pedestrian interference. Despite these challenges, the AI-replicated navigation cells performed well, with DR trajectories closely matching NeuroSLAM and maintaining consistent accuracy. This confirms that the nonlinear motions did not seriously affect the encoding and decoding functions of the AI-replicated navigation cells, and their robustness against instantaneous changes was verified experimentally.

Next, we present the neurodynamics envelope activity patterns obtained by replicating CANN-modeled HDC and GC using HDCN and GCN. Figure 8 shows the changes in the neurodynamics of the AI-replicated HDCN during changes in altitude. When the system's altitude state changes, as shown in panels (a-1) to (a-4) and (c-1) to (c-4), the activity bump state decoded from the HDCN undergoes corresponding smooth movement along the altitude dimension. When the bump exceeds the boundary range, the HDCN can excite neighboring neurons on the opposite side, allowing the bump state to transition continuously and smoothly, fully replicating the activity continuity of CANN. When the system altitude remains constant, the bump state decoded from the HDCN remains stable in altitude and continues to move along the yaw dimension, as demonstrated in the groups (b-1) to (b-4) in Figure 8.

Figure 9 shows the bump activation patterns decoded from the GCN under different altitude change modes. As shown in panels (a-1) to (a-4) and (c-1) to (c-4), when the altitude changes, the bump moves according to the trend of altitude change. The bump, similar to the HDC, activates neighboring neurons on the opposite side, exhibiting the continuous, stable, and periodic activity pattern characteristic of CANN, when the cumulative change exceeds the representational range. When the altitude is unchanged, as shown in panels (b-1) to (b-4) in Figure 9, the bump activity decoded from the GCN remains stable on the Z axis and moves along the X axis, crossing boundaries.

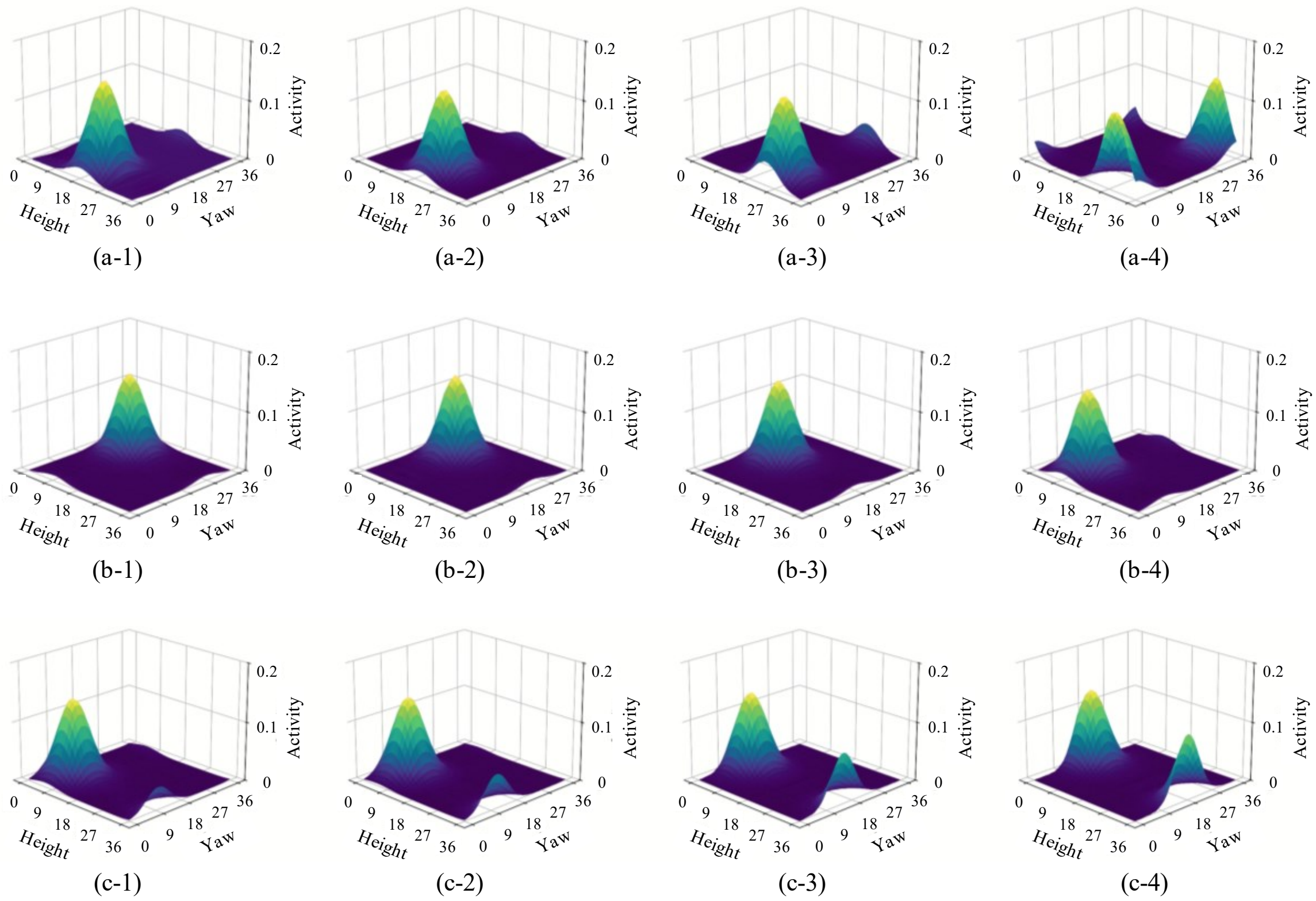

**Fig. 8.** Neural activity patterns decoded from the HDCN. (a-1) to (a-4): rising height. (b-1) to (b-4): unchanged height. (c-1) to (c-4): falling height.

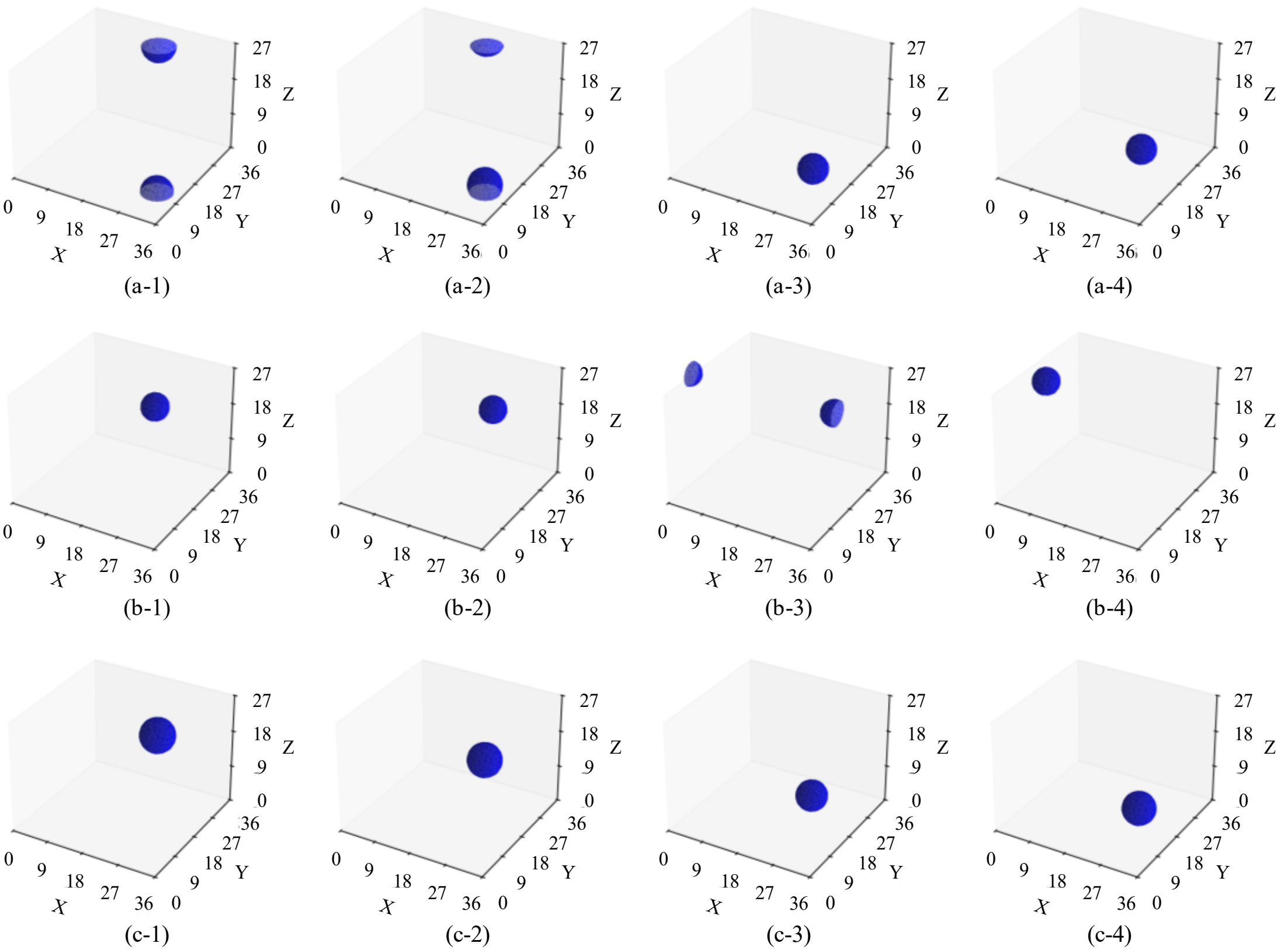

**Fig. 9.** Neural activity patterns decoded from the GCN. (a-1) to (a-4): rising height. (b-1) to (b-4): unchanged height. (c-1) to (c-4): falling height.

From the above analysis, we can qualitatively conclude that the learning-based approach for replicating CANN's neurodynamics proposed in this paper is successful. However, the specific CANN modeling method adopted here is not the most advanced. For instance, many novel CANN modeling methods have been reported in recent years, such as [28][29]. Therefore, this work has room for further extension, and we believe it is worthwhile to explore its application to more comprehensive CANN modeling methods. Furthermore, the lightweight nature of this work renders its quantization and acceleration on edge AI chips, and even deployment on neuromorphic hardware via ANN2SNN transformation [30], as promising directions for future exploration.

## V. Discussion and Analysis

The traditional navigation pipeline is often described as "perception-planning/decision-control". Building capabilities at the perception level is the foundation of navigation and localization. Consequently, within the emerging field of BIN, there are numerous advances themed around brain-inspired SLAM [3][4].

❑ Why PI is important for BIN?

As analyzed in Section II, simulating the PI capability of the entorhinal cortex is a core component in almost all existing brain-inspired SLAM systems based on CANNs. Essentially, the core logic of PI involves integrating internal and external cues captured by the BIN system by simulating the collaborative logic between GCs and other navigation cells, thereby forming spatial experiences. How to effectively utilize the spatial experiences generated by PI is the primary concern for BIN systems.

Therefore, simulating PI capability is a core mechanism of brain-inspired SLAM but is not limited to the SLAM topic. For instance, Edwards et al. [31][32] utilized CANNs to simulate the PI capability of GCs for vector-based navigation tasks. Yang et al. [14] directly implemented a DR task with a CANN-based multi-scale GC model to build PI capability, without involving LCD as in SLAM. The method developed in this study also focuses on realizing PI capability, aiming to achieve the DR capability for BIN systems directly.

After clarifying the essence of simulating the brain's PI capability, we need to consider its modeling effectiveness and computational efficiency. The former aims to enable the BIN system to achieve a more comprehensive replication of the brain's PI capability, while the latter is pursued to promote the practicality of brain-inspired PI models. Among existing studies on simulating brain-inspired PI capability, most advances belong to the former category, whereas the work presented herein belongs to the latter.

❑ Discussion on time-cost analysis

Workstations have ample computing power, showing less improvements compared to edge devices like Raspberry Pi, but the nearly 20% efficiency gain in this work is still very valuable for long-endurance navigation, especially for field unmanned systems with industrial computers. Moreover, native CANNs, which rely on iterative neurodynamics simulation through differential equations, can strain resource-constrained devices and cause computational bottlenecks. Therefore, replacing them with lightweight ANNs is practical for edge devices.

Moreover, it is worth noting that, for the current version, this work indeed outperforms the native CANN-based PI implementation in computational efficiency. However, during memory monitoring on the Raspberry Pi, we found that the PyTorch and TensorFlow libraries required for ANN inference occupy an additional 500MB of memory during algorithm initialization. Specifically, in the NeuroSLAM system, the memory usage for the CANN-based PI algorithm increased from 64.37MB to 84.14MB from initialization to operation end. In contrast, our work's memory usage increased from 514.45MB to 530.04MB.

In the future, we plan to enhance this work using model compression, quantization, and Tiny Machine Learning (TinyML) techniques. This will improve efficiency through matrix calculations (binary file) and eliminate the large library dependencies required by native PyTorch or TensorFlow frameworks [33]. It will also reduce inference overhead for ANN components and even enable deployment on resource-constrained Microcontroller Units (MCUs). We also plan to explore ANN2SNN solutions in future work to develop neuromorphic edge solutions.

❑ Discussion on the integration potential with LCD

Section IV has conducted benchmark tests after implementing the approach of replicating CANN's neurodynamics using lightweight ANNs. Apart from the PI capability, the main difference between this work and the NeuroSLAM system lies in the LCD. In Experiment 1, the open-source data from the NeuroSLAM work was used. Debug its original code, we reveal that, under its default parameters, the influence of its LCD capability on the construction of the experience map is relatively weak. As for Experiment 2, we likewise noticed this issue on our self-collected dataset, that is, the LCD component of NeuroSLAM did not perform the expected map correction.

After meticulous step-by-step code debugging, we suspect the reason might be that the parameter configuration for this dataset was not optimally tuned, leading to short durations for successful visual template matching (triggering effective energy injection) and a very small intervention impact during continuous energy injection. Therefore, on this dataset, the relaxation and association process within NeuroSLAM's experience map played the dominant role in modifying and adjusting the spatial experience. Consequently, the DR trajectories from both methods almost completely overlapped, enabling this work to achieve results on par with NeuroSLAM, despite lacking LCD. Interested readers can verify it themselves using our open-source Python implementation of NeuroSLAM. URL: https://github.com/BINUCOE/OpenNeuroSLAM.

Given the differences in computational paradigms between ANNs and native CANNs, we need to highlight a limitation of the current work: ANN weights are fixed after training and do not change during inference, unlike the native CANNs, which allows "energy injection" for state adjustment (though highly influenced by preset parameters). We plan to address this in future work by exploring dynamic neural network technologies to enable adjustable weights [34], allowing similar "energy injection" for state adjustment. Additionally, we have provided a supplementary explanation in the open-source repository of this work on how to simply integrate it with LCD based on visual template matching. For details, see: https://github.com/BINUCOE/CANN2ANN_for_PI.

In fact, PI merely acts like an accumulator, aggregating self-motion cues according to the spatial experience connection relationships stored in each spatial experience node. The resulting trajectory error level relies more on the correction of the relaxation association algorithm. In fact, we are further exploring and analyzing the introduction of methods such as geometric consistency constraints into the relaxation association process of spatial experience, and combining them with suitable LCD methods for joint optimization, to further improve the accuracy of DR under brain-inspired PI framework.

## VI. Conclusion

This paper presents an efficient brain-inspired PI method by replicating the neurodynamics of CANNs using lightweight ANNs. This method avoids computational redundancy in the RatSLAM-derived PI modeling, enhancing efficiency. It achieved roughly 17.5% efficiency gains on general-purpose devices and 40~50% on edge devices compared to the well-known NeuroSLAM. Overall, the benchmark tests consistently reported performance equivalent to NeuroSLAM. Moreover, on this basis, the computational efficiency improvement brought by this work also demonstrates its value, particularly for applications on edge devices. Thereby, it promotes the practicality of BIN technology.

In the future, we hope to go beyond the existing ideas and replicate the neurodynamic characteristics of navigation cells modeled by more advanced CANN models. Moreover, building PI capability is a core issue for brain-inspired SLAM systems, but this paper has currently only explored the brain-inspired PI method itself, achieving DR capability. The future research directions encompass integrating this work into a complete SLAM pipeline with effective loop closure detection strategies, and investigating its deployment potential on edge AI accelerators and neuromorphic chips. In summary, this paper provides a feasible path to realizing a scalable and efficient BIN system, with the potential for practicality and continuous expansion.

## Reference

[1] Bai, Y., Shao, S., et al., "A review of brain-inspired cognition and navigation technology for mobile robots," Cyborg and Bionic Systems, vol. 5, p. 0128, 2024.

[2] McNaughton, B.L., Battaglia, F.P., et al., "Path integration and the neural basis of the 'cognitive map'," Nature Reviews Neuroscience, vol. 7, no. 8, pp. 663-678, 2006.

[3] He, X., Meng, X., et al., "A comprehensive review of brain-inspired navigation," ACM Computing Surveys, vol. 58, no. 8, p. 200, 2026.

[4] He, X., Meng, X., et al., "A preliminary exploration of the differences and conjunction of traditional PNT and brain-inspired PNT," ArXiv, abs/2510.16771, 2025.

[5] Milford, M., Wiles, J., et al., "Solving navigational uncertainty using grid cells on robots," PLoS Computational Biology, vol. 6, no. 11, p. e1000995, 2010.

[6] Ball, D., Heath, S., et al. OpenRatSLAM: an open source brain-based SLAM system," Autonomous Robots, vol. 34, no. 3, pp. 149-176, 2013.

[7] Yuan, M., Tian, B., et al., "An entorhinal-hippocampal model for simultaneous cognitive map building," AAAI Conference on Artificial Intelligence (AAAI), pp. 586-592, 2015.

[8] Zeng, T., Tang, F., et al., "NeuroBayesSLAM: neurobiologically inspired Bayesian integration of multisensory information for robot navigation," Neural Networks, vol. 126, pp. 21-35, 2020.

[9] Lu, T., Wang, Z., et al., "Hybrid-NeuroSLAM: a neurobiologically inspired hybrid visual-inertial SLAM method for large scale environment," IEEE Robotics and Automation Letters, vol. 10, no. 7, pp. 7484-7491, 2025.

[10] Michael M., Wyeth, G.F., et al., "RatSLAM: a hippocampal model for simultaneous localization and mapping," IEEE International Conference on Robotics and Automation (ICRA). pp. 403-408, 2004.

[11] Yu, F., Shang, J., et al., "NeuroSLAM: a brain-inspired SLAM system for 3D environments," Biological Cybernetics, vol. 113, no. 5-6, pp. 515-545, 2019.

[12] Shen, D., Liu, G., et al., "ORB-NeuroSLAM: a brain-inspired 3-D SLAM system based on ORB features," IEEE Internet of Things Journal, vol. 11, no. 7, pp. 12408-12418, 2024.

[13] Chen, Y., Xiong, Z., et al., "Brain-inspired multisensor navigation information fusion model based on spatial representation cells," IEEE Sensors Journal, vol. 24, no. 11, pp. 18122-18132, 2025.

[14] Yang, C., Xiong, Z., et al., "A path integration approach based on multiscale grid cells for large-scale navigation," IEEE Transactions on Cognitive and Developmental Systems, vol. 14, no. 3, pp. 1009-1020, 2022.

[15] Yang, C., Xiong, Z., et al., "Brain-inspired multimodal navigation with multiscale hippocampal-entorhinal neural network," IEEE Transactions on Instrumentation and Measurement, vol. 73, p. 8507217, 2024.

[16] Liu, X., Chen, L., et al., "A neuro-inspired positioning system integrating MEMS sensors and DTMB signals," IEEE Transactions on Broadcasting, vol. 69, no. 3, pp. 823-831, 2023.

[17] Zhang, Y., Trappenberg, T., et al., "Continuous attractor neural networks: candidate of a canonical model for neural information representation," F1000Research, vol. 5, p. 156, 2016.

[18] Finkelstein, A., Las, L., et al., "3-D maps and compasses in the brain," Annual Review of Neuroscience, vol. 39, pp. 171-196, 2016.

[19] Menezes, M.C., Muñoz, M.E., et al., "A Multisession SLAM Approach for RatSLAM," Journal of Intelligent & Robotic Systems, vol. 108, no. 4, pp. 61, 2023.

[20] Zeng, T., Si, B., "Cognitive mapping based on conjunctive representations of space and movement," Frontiers in Neurorobotics, vol. 11, pp. 61, 2017.

[21] Silveira, L., Guth, F., et al., "An open-source bio-inspired solution to underwater SLAM," IFAC-PapersOnLine, vol. 48, no. 2, pp. 212-217, 2015.

[22] Milford, M., Wyeth, G.F., "Mapping a suburb with a single camera using a biologically inspired SLAM system," IEEE Transactions on Robotics, vol. 24, no. 25, pp. 1038-1053, 2008.

[23] Steckel, J., Peremans, H., "BatSLAM: simultaneous localization and mapping using biomimetic sonar," PLoS ONE, vol. 8, no. 1, p. e54076, 2013.

[24] Yuan, J., Guo, W., et al., "A bionic spatial cognition model and method for robots based on the hippocampus mechanism," Frontiers in Neurorobotics, vol. 15, p. 769829, 2022.

[25] Wang, C., Zhang, T., et al., "BrainPy, a flexible, integrative, efficient, and extensible framework for general-purpose brain dynamics programming," eLife, vol. 12, p. e86365, 2023.

[26] Cueva C., Wei X., "Emergence of grid-like representations by training recurrent neural networks to perform spatial localization," International Conference on Learning Representations (ICLR), pp. 1512-1530, 2018.

[27] Lin J., Wang Y., et al., "Path integration based on a recurrent spiking neural network," Electronic Research Archive, vol. 34, no. 1, pp. 1-30, 2026.

[28] Kymn, C.J., Mazelet, S., et al., "Binding in hippocampal-entorhinal circuits enables compositionality in cognitive maps," International Conference on Neural Information Processing Systems (NeurIPS), 2024.

[29] Dong, X., Ji, Z., et al., "Adaptation accelerating sampling-based Bayesian inference in attractor neural networks," International Conference on Neural Information Processing Systems (NeurIPS), 2024.

[30] Fang, W., Chen Y., et al., "SpikingJelly: an open-source machine learning infrastructure platform for spike-based intelligence," Science

Advances, vol. 9, no. 40, p. eadi1480, 2023.
[31] Edvardsen, V., “Goal-directed navigation based on path integration and decoding of grid cells in an artificial neural network,” Natural Computing, vol. 18, no. 1, pp. 13-27, 2019.
[32] Edvardsen, V., Bicanski, A., et al., “Navigating with grid and place cells in cluttered environments,” Hippocampus, vol. 30, no. 3, pp. 220-232, 2019.
[33] Wu, Y., He, X., et al., “Self-attention-assisted TinyML with effective representation for UWB NLOS identification,” IEEE Internet of Things Journal, vol. 11, no. 15, pp. 25471-25480, 2024.
[34] He, X., Meng, X., et al., “Enhance UWB performance using dynamic AI-enabled approach,” IEEE Internet of Things Journal, vol. 12, no. 24, pp. 52474-52487, 2025.

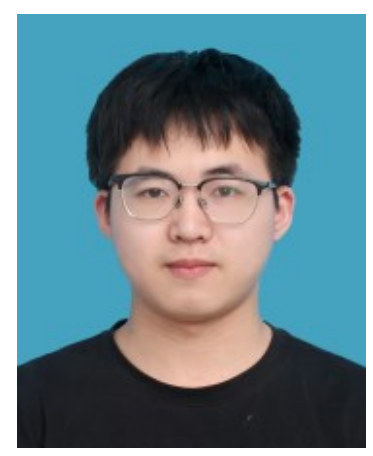

**Zhangyu Ge**, is with the School of Instrument Science and Engineering, Southeast University, Nanjing, China (e-mail: 220233603@seu.edu.cn). He is pursuing a M.S. degree in instrumentation science and technology with the School of Instrument Science and Engineering of Southeast University. His research interests focus on brain-inspired navigation, SLAM.

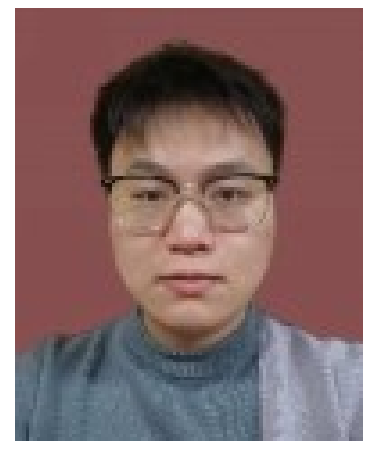

**Xu He**, (*Student Member, IEEE*), is with the School of Instrument Science and Engineering, Southeast University, Nanjing, China (e-mail: hexu@seu.edu.cn). He is pursuing a Ph.D. degree in instrumentation science and technology with the School of Instrument Science and Engineering of Southeast University. He has authored 8 JCR Q1/Q1 TOP papers as the first/co-first author. His research interests focus on brain-inspired PNT. He is a student member of IEEE.

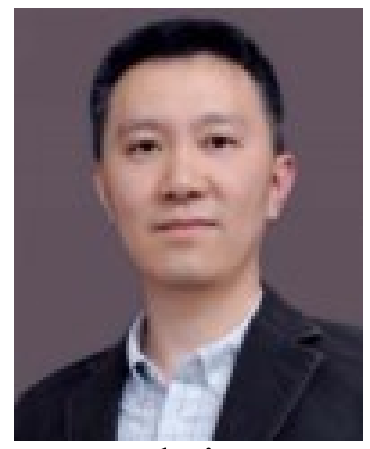

**Lingfei Mo,** (*Member, IEEE*), is with the School of Instrument Science and Engineering, Southeast University, Nanjing, China (e-mail: lfmo@seu.edu.cn). He is an Associate Professor at the School of Instrument Science and Engineering, Southeast University, Nanjing, China. His research interests focus on IoT, AI, and brain-like computing. He has authored over 100 technical publications in standard journals and conferences. Dr. Mo is also a member of the IEEE and an Associate Editor of the IEEE Journal of Radio Frequency Identification (RFID).

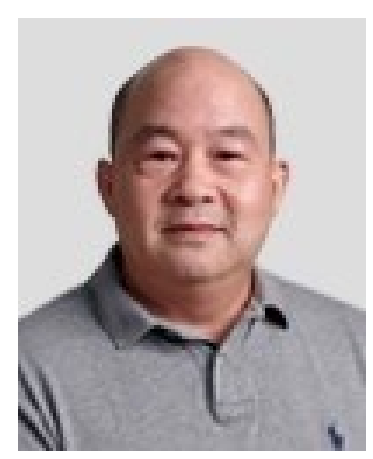

**Xiaolin Meng,** is with the School of Instrument Science and Engineering, Southeast University, Nanjing, China, as Chair Professor of Intelligent Mobility (e-mail: xiaolin_meng@seu.edu.cn). He was a Professor of Intelligent Mobility and a Theme Leader of PNT technologies with the University of Nottingham, Nottingham, U.K., until May 2020. He has authored more than 400 publications and undertaken a research portfolio of more than £20 m. His research interests mainly include intelligent transportation systems and smart cities, CAV, structural health monitoring of large infrastructure, and precision agriculture/ livestock farming.

Dr. Meng is the first-ever Professor of Intelligent Mobility in the UK. He has been selected by the UK’s Research Council as Academic Fellow, and is a Fellow of the RIN and IET, and a Chartered Engineer (CEng) of the UK’s Engineering Council. He is the Executive Editor, Associate Editor, Guest Editor, Editor and Reviewer for over 30 important international journals. His research achievements have been featured in 16 special reports by Science and Technology Daily and BBC. He has served as the chairman of the Institute of Navigation (ION) session and received the ION Best Paper Award and the 2019 UK Engineer “Collaborate to Innovate” Award. He is the founder of the Sino-UK Geospatial Engineering Center and also the founder and chief scientist of UbiPOS UK LTD.

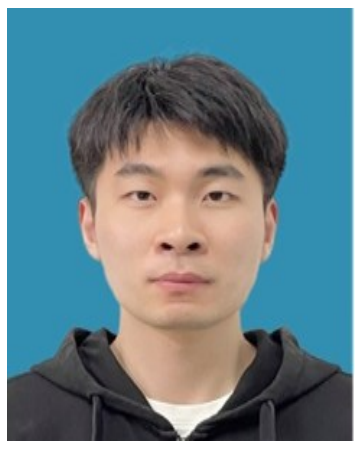

**Wenxuan Yin**, (*Student Member, IEEE*), is with the School of Instrument Science and Engineering, Southeast University, Nanjing, China (e-mail: wenxuan_yin@seu.edu.cn). He is pursuing a Ph.D. degree with the School of Instrument Science and Engineering of Southeast University. He has authored 3 SCI papers. His research interests focus on brain-inspired PNT and intelligent control.

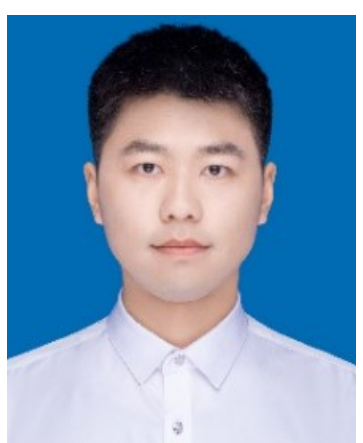

**Youdong Zhang,** (*Student Member, IEEE*), is with the School of Instrument Science and Technology, Southeast University, Nanjing, China (e-mail: ydzhang@seu.edu.cn). He is pursuing a Ph.D. degree in instrumentation science and technology with the School of Instrument Science and Engineering of Southeast University. He has authored/co-authored 5 JCR Q1/Q1 TOP papers. His research interests focus on brain-inspired PNT and brain-inspired intelligence.

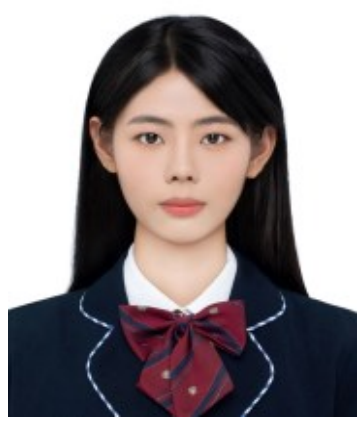

**Lansong Jiang**, is with the School of Instrument Science and Technology, Southeast University, Nanjing, China (e-mail: 220243706@seu.edu.cn). She is pursuing a M.S. degree with the School of Instrument Science and Engineering of Southeast University. Her research interests focus on vehicle-to-infrastructure cooperative localization.

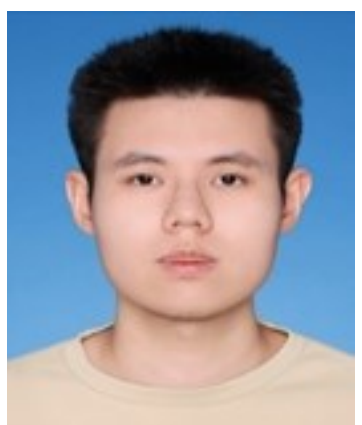

**Fengyuan Liu,** received the B.S. degree in Intelligent Sensing Engineering from Southeast University, Nanjing, China, in 2024 (e-mail: 230268565@seu.edu.cn), where he is currently pursuing the Ph.D. degree with the School of Instrument Science and Engineering. His current research interests include C-V2X collaboration and GNSS PNT.